\documentclass[11pt,letterpaper]{article}
\usepackage{emnlp2017}
\usepackage{times}
\usepackage{latexsym}

\usepackage{amsfonts}
\usepackage{graphicx}
\usepackage{epstopdf}
\usepackage[boxed,linesnumbered,lined]{algorithm2e}
\usepackage{algpseudocode}
\usepackage{multirow}
\usepackage{bm}

\emnlpfinalcopy

\def\emnlppaperid{37}

\title{THUMT: An Open Source Toolkit for Neural Machine Translation}

\author{Jiacheng Zhang$^\dagger$, Yanzhuo Ding$^\dagger$, Shiqi Shen$^\dagger$, Yong Cheng$^\#$, \\{\bf Maosong Sun$^{\dagger\ddagger}$,  Huanbo Luan$^\dagger$ and Yang Liu$^{\dagger\ddagger}$\thanks{\ \ Corresponding author: Yang Liu.} }\\
    $^\dagger$State Key Laboratory of Intelligent Technology and Systems  \\
    Tsinghua National Laboratory for Information Science and Technology \\
    Department of Computer Science and Technology, Tsinghua University, Beijing, China \\
    $^\ddagger$Jiangsu Collaborative Innovation Center for Language Competence, Jiangsu, China\\
    $^\#$Institute for Interdisciplinary Information Sciences, Tsinghua University, Beijing, China
    }

\date{}

\begin{document}

\maketitle

\begin{abstract}

This paper introduces THUMT, an open-source toolkit for neural machine translation (NMT) developed by the Natural Language Processing Group at Tsinghua University. THUMT implements the standard attention-based encoder-decoder framework on top of Theano and supports three training criteria: maximum likelihood estimation, minimum risk training, and semi-supervised training. It features a visualization tool for displaying the relevance between hidden states in neural networks and contextual words, which helps to analyze the internal workings of NMT. Experiments on Chinese-English datasets show that THUMT using minimum risk training significantly outperforms GroundHog, a state-of-the-art toolkit for NMT.
 
\end{abstract}

\section{Introduction}

End-to-end neural machine translation (NMT) \cite{Sutskever:14,Bahdanau:15} has gained increasing popularity in the machine translation community. Capable of capturing long-distance dependencies with gating \cite{Hochreiter:97,Cho:14} and attention \cite{Bahdanau:15} mechanisms, NMT has proven to outperform conventional statistical machine translation systematically across a variety of language pairs \cite{Junczys-Dowmunt:16}. 

This paper introduces THUMT, an open-source toolkit developed by the Tsinghua Natural Language Processing Group. On top of Theano \cite{bergstrao:10}, THUMT implements the standard attention-based encoder-decoder framework for NMT \cite{Bahdanau:15}. It supports three training criteria: maximum likelihood estimation \cite{Bahdanau:15}, minimum risk training \cite{Shen:16}, and semi-supervised training \cite{Cheng:16}. To facilitate the analysis of the translation process in NMT, THUMT also provides a visualization tool that calculates the relevance between hidden layers of neural networks and contextual words. We compare THUMT with the state-of-the-art open-source toolkit GroundHog \cite{Bahdanau:15} and achieve significant improvements on Chinese-English translation tasks by introducing new training criteria and optimizers.

\begin{figure*}[!t]
	\begin{center}
		\includegraphics[width=0.95\textwidth]{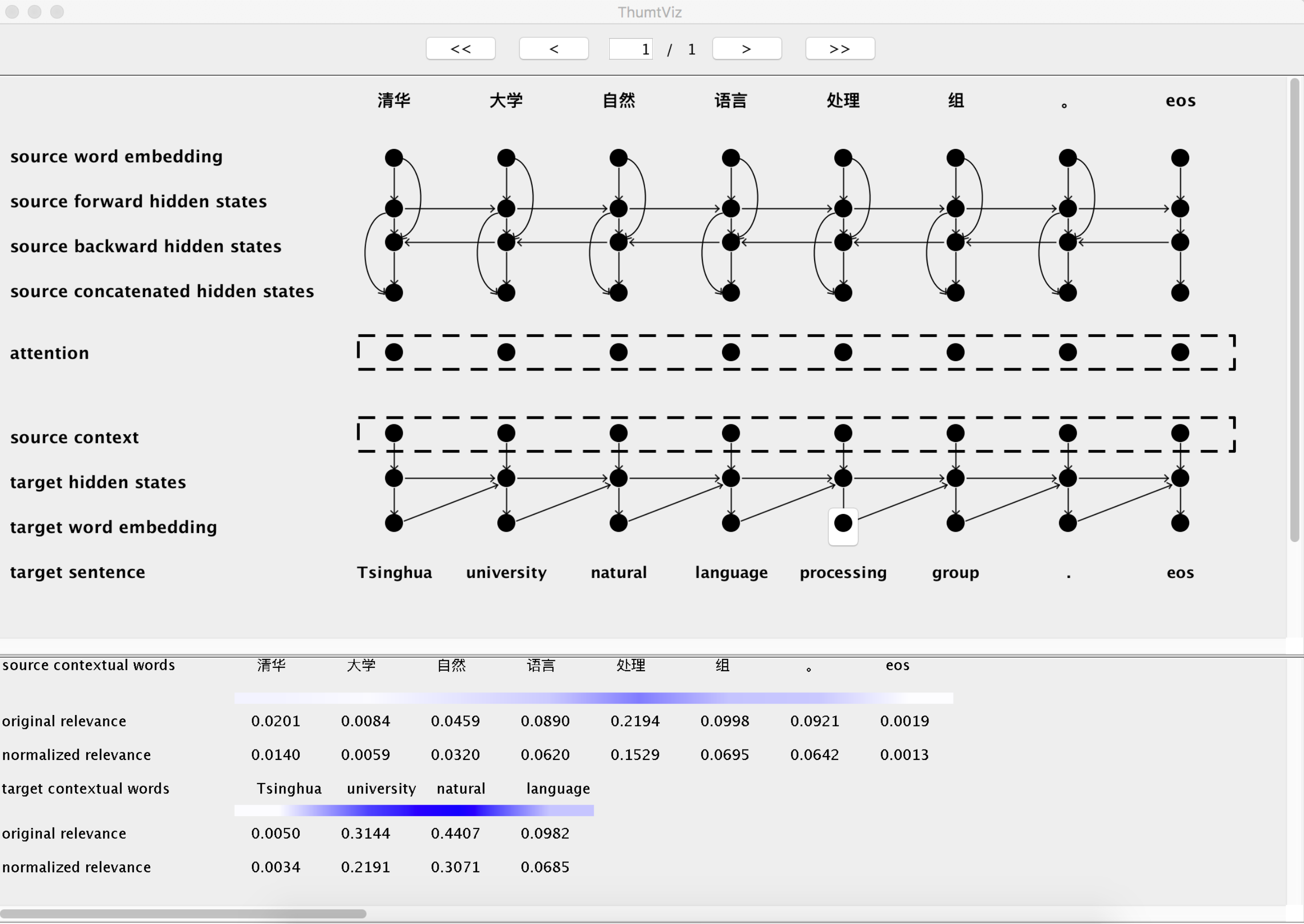}
	\end{center}
	\caption{Visualizing neural machine translation.}
	\label{fig:viz}
\end{figure*}

\section{The Toolkit}
\label{sec:Toolkit}

%The toolkit mainly consists of a trainer and a translator. To train a new machine translation model, all we need is to prepare the parallel corpus. Then the training process can be started with a single command. Once the training process is finished, translation can be done with the translation model and the easy-to-use translator. 

%We will give a brief introduction on the components and features provided by the toolkit.

\subsection{Model}

THUMT implements the standard attention-based NMT model \cite{Bahdanau:15} on top of Theano \cite{bergstrao:10}.  Please refer to \cite{Bahdanau:15} for more details.

\subsection{Training Criteria}

THUMT supports three training criteria: 

\begin{enumerate}
    \item{\bf Maximum likelihood estimation} (MLE) \cite{Bahdanau:15}: the default training criterion in THUMT, which aims to find a set of model parameters that maximizes the likelihood of training data.
    \item{\bf Minimum Risk Training} (MRT) \cite{Shen:16}: the recommended training criterion in THUMT, which aims to find a set of model parameters that minimizes the risk (i.e., expected loss measured by evaluation metrics) on training data. In THUMT, MLE is often used to initialize MRT. In other words, the model trained with respect to MLE serves as the initial model of MRT.
    \item{\bf Semi-supervised Training} (SST) \cite{Cheng:16}: the recommended training criterion for low-resource language translation. SST is capable of exploiting abundant monolingual corpora to train source-to-target and target-to-source translation models jointly. MLE is also used to initialize SST.
\end{enumerate}

\subsection{Optimization}

Optimization plays a crucial role in NMT and directly influences the training time and translation quality. THUMT supports three optimizers:

\begin{enumerate}
    \item{\bf SGD}: standard stochastic gradient descent with fixed learning rate. 
    \item{\bf Adadelta} \cite{Zeiler:12}: dynamically adapting learning rate over time according to history.
    \item{\bf Adam} \cite{kingma:15}:  computing individual learning rate for different parameters. THUMT uses a modified version of Adam to address the NaN problem.
\end{enumerate}

%\begin{figure}[!t]
%	\begin{center}
%		\includegraphics[width=0.48\textwidth]{figure/speed.eps}
%	\end{center}
%	\caption{Comparison of training time.}
%	\label{fig:speed}
%\end{figure}

\begin{table*}[!t]
\centering

\begin{tabular}{c|c|c||c|ccccc|c}
Toolkit & Criterion & Optimizer & MT02 & MT03 & MT04 & MT05 & MT06 & MT08 & All \\
\hline \hline
\multirow{2}{*}{GroundHog} & \multirow{2}{*}{MLE}  & AdaDelta & 33.14 & 30.22 & 32.15 & 30.15 & 28.94 & 22.44 & 29.00 \\
&  & Adam & 35.30 & 32.31 & 34.83 & 31.66 & 31.21 & 23.31 & 30.96 \\
\hline \hline
\multirow{4}{*}{\textproc{THUMT}} & \multirow{2}{*}{MLE} & AdaDelta & 33.45 & 30.93 & 32.57 & 29.86 & 29.03 & 21.85 & 29.11 \\
&  & Adam & 35.82 & 33.07 & 35.22 & 32.47 & 31.19 & 23.58 & 31.38 \\
\cline{2-10}
& \multirow{2}{*}{MRT} & AdaDelta & 37.39 & 34.50 & 36.95 & 34.46 & 33.24 & 25.36 & 33.20 \\
&  & Adam & 40.67 & 37.41 & 39.87 & 37.45 & 36.80 & 27.98 & 36.22  \\
\end{tabular}
\caption{Comparison between GroundHog and THUMT.} \label{table:main}
\end{table*}

\begin{table*}[!t]
\centering

\begin{tabular}{c|c||c|ccccc|c}
Criterion & Direction & MT02 & MT03 & MT04 & MT05 & MT06 & MT08 & All \\
\hline \hline
\multirow{2}{*}{MLE} & zh $\rightarrow$ en & 35.82 & 33.07 & 35.22 & 32.47 & 31.19 & 23.58 & 31.38 \\
& en $\rightarrow$ zh & 20.96 & 15.65 & 16.70 &  14.04 & 15.31 & 12.69 & 15.07 \\
\hline
\multirow{2}{*}{SST} & zh $\rightarrow$ en & 40.04 & 37.51 & 39.72 & 37.73 & 36.00 & 28.71 & 36.22 \\
& en $\rightarrow$ zh & 24.13 & 18.82  & 18.54  & 16.28  & 18.21  & 14.00  & 17.50   \\
\end{tabular}
\caption{Comparison between MLE and SST.
} \label{table:semi}
\end{table*}

\begin{table}[!t]
\centering

\begin{tabular}{l|c||c|c}
Criterion & Optimizer & w/o & w/ \\
\hline \hline
\multirow{2}{*}{MLE} & AdaDelta & 29.11 & 29.66 \\
 & Adam & 31.38 & 31.94 \\
 \hline
\multirow{2}{*}{MRT}& AdaDelta & 33.20 & 33.76\\
 & Adam & 36.22 & 36.92 \\

\end{tabular}
\caption{Effect of replacing unknown words.} \label{table:unkreplace}
\end{table}

\subsection{Visualization}

Although NMT achieves state-of-the-art translation performance, it is hard to understand how it works because all internal information is represented as real-valued vectors or matrices. To address this problem, THUMT features a visualization tool to use layer-wise relevance propagation (LRP) \cite{Bach:15} to visualize and interpret neural machine translation \cite{Ding:17}. 

Figure \ref{fig:viz} shows an example. Given a source sentence, a target sentence, and a trained model, THUMT displays the entire attention-based neural network. Clicking on a node of the network (e.g., the output node of ``processing''), the relevance values of relevant source and target contextual words are shown in the bottom area. This is helpful for analyzing the internal workings of NMT. Please refer to \cite{Ding:17} for more details.

\subsection{Replacing Unknown Words}

We follow \citet{Luong:15} to address unknown words. In our implementation, we use FastAlign \cite{Dyer:13} to generate bilingual dictionaries with alignment probabilities.

%\subsection{User Interface}

%We optimize user interface of the toolkit to bring convenience to users. The training process can be started in a single command as long as parallel corpus is prepared. The toolkit automatically validates the model periodically and saves validation results in a log file, enabling users to check the training status with ease. After training, All information required by the translation process is stored in a single model file. Translation can be done without referring to any file other than the model file.

%With the quick guide included in the toolkit, users can build the translation system without knowing NMT. The quick guide describes the steps of training and translating in detail. The parameters related to the translation model are kept transparent to non-research users.

%We provide full documentation for research users, in which we explain each component and each parameter in detail. We arrange the components in an extensible framework, allowing research users to understand the model without difficulty.

\section{Experiments}
\label{sec:Experiments}

\subsection{Setup}

We evaluate THUMT on the Chinese-English translation task. The training set contains 1.25M sentence pairs with 27.9M Chinese words and 34.5M English words. For SST,  the Chinese
monolingual corpus contains 18.75M sentences with 451.94M words. The English corpus contains
22.32M sentences with 399.83M words. The vocabulary sizes of Chinese and English are 0.97M
and 1.34M, respectively. The validation set is NIST 2002 dataset and the test sets are NIST 2003, 2004, 2005, 2006 and 2008 datasets. The evaluation metric is case-insensitive BLEU \cite{papineni:02} score. Our baseline system is GroundHog \cite{Bahdanau:15}, a state-of-the-art open-source NMT toolkit.

We use the same setting of hyper-parameters for both GroundHog and THUMT. The vocabulary size is set to 30K. We set word embedding dimension to 620 for both languages. The dimension of hidden layers is set to1000. In training, we set the mini-batch size to 80. In decoding, we set the beam size to 10. During training, we set the step sizes of Adam optimizer to 0.0005 for MLE,  0.00001 for MRT, and 0.00005 for SST, respectively.  We train NMT models on a single GPU device Tesla M40.

%\begin{figure}[!t]
%	\begin{center}
%		\includegraphics[width=0.48\textwidth]{figure/target.pdf}
%	\end{center}
%	\caption{Visualizing source hidden states for target content word ``struggle'' .}
%	\label{fig:target}
%\end{figure}

\subsection{Results}

\begin{table}[!t]
\centering
\begin{tabular}{c|c||r|r}
 Criterion & Optimizer & Iteration & Time \\
\hline \hline
 \multirow{2}{*}{MLE} & AdaDelta & 200K & 55.9h \\
 & Adam &  36K & 10.1h \\
 \hline
 \multirow{2}{*}{MRT} & AdaDelta &  680K & 389.5h  \\
  & Adam & 118K & 72.1h  \\
  \hline
 SST & Adam & 12K & 61.8h  \\
\end{tabular}
\caption{Comparison of training time between MLE, MRT, and SST.} \label{table:speed}
\end{table}

Table \ref{table:main} shows the BLEU scores obtained by GroundHog and THUMT using different training criteria and optimizers. Experimental results show that the translation performance of THUMT is comparable to \textproc{Groundhog} using MLE.  Due to the capability to include evaluation metrics in during, MRT obtain significant improvements over MLE. \footnote{Currently, THUMT only supports single GPUs. In the original paper \cite{Shen:16}, MRT actually runs on multiple GPUs and enables more samples to fit in the memory. We will release a new version that supports multiple GPUs soon.} Another finding is that Adam leads to consistent and significant improvements over AdaDelta.

Table \ref{table:semi} shows the effect of semi-supervised training. It is clear that exploiting monolingual corpora helps to improve translation quality for both directions. 

Table \ref{table:unkreplace} shows that replacing unknown words leads to consistent improvements for all training criteria and optimizers.

 Table \ref{table:speed} compares the training time between MLE, MRT, and SST. While ``MLE + AdaDelta'' requires 200K iterations and 55.9 hours to converge, ``MLE + Adam'' only needs 36K iterations and 10.1 hours. The training time of MRT is much longer than MLE.

%\begin{figure}[!t]
%	\begin{center}
%		\includegraphics[width=0.475\textwidth]{figure/error.pdf}
%	\end{center}
%	\caption{Analyzing translation error: translation omission.}
%	\label{fig:error}
%\end{figure}

\section{Conclusion}

We have introduced a new open source toolkit for NMT that supports two new training criteria: minimum risk training \cite{Shen:16} and semi-supervised training \cite{Cheng:16}. While minimum risk training proves to improve over standard maximum likelihood estimation substantially, semi-supervised training is capable of exploiting monolingual corpora to improve low-resource translation. The toolkit also features a visualization tool for analyzing the translation process of THUMT. The toolkit is freely available at \url{http://thumt.thunlp.org}.

\section*{Acknowledgements}
This research is supported by the National Natural Science Foundation of China (No. 61522204).

\bibliography{emnlp2017}
\bibliographystyle{emnlp_natbib}

\end{document}